\documentclass[pdflatex,sn-basic]{sn-jnl}%
\usepackage{swp-patch}
\usepackage{graphicx}
\usepackage{multirow}
\usepackage{amsmath,amssymb,amsfonts}
\usepackage{amsthm}
\usepackage{mathrsfs}
\usepackage[title]{appendix}
\usepackage{xcolor}
\usepackage{textcomp}
\usepackage{manyfoot}
\usepackage{booktabs}
\usepackage{algorithm}
\usepackage{algorithmicx}
\usepackage{algpseudocode}
\usepackage{listings}
\usepackage{amsmath}
\usepackage{amsfonts}
\usepackage{amssymb}%
\providecommand{\U}[1]{\protect\rule{.1in}{.1in}}
\providecommand{\U}[1]{\protect\rule{.1in}{.1in}}
\theoremstyle{thmstyleone}

\theoremstyle{thmstyletwo}

\theoremstyle{thmstylethree}

\affiliation{\aff{1}Department of Engineering Mechanics, Tsinghua University, Beijing, China}
\begin{document}

\title{Duality-based Mode Operations and Pyramid Multilayer Mapping for Rhetorical Modes}
\author{Zi-Niu WU\aff{1}*\corresp{E-mail: ziniuwu@tsinghua.edu.cn}}
\maketitle

\abstract{Rhetorical modes are useful in both academic and non-academic writing, and can be subjects
to be studied within linguistic research and computational modeling. Establishing a conceptual bridge
among these domains could enable each to benefit from the others. This paper proposes duality-based
mode operations (split-unite, forward-backward, expansion-reduction and orthogonal dualities) to
expand the set of rhetorical modes, introducing generated modes like combination and generalization,
thereby enhancing epistemic diversity across multiple applications. It further presents a pyramid multilayer
mapping framework (e.g., three layers from the rhetorical model layer, to cognitive layer, and to epistemic layers) that reduces
the resulting cognitive complexity. The degrees of expressive diversity and complexity reduction are
quantified through binomial combinatorics and Shannon entropy analysis. A Marginal Rhetorical Bit (MRB)
is identified, permitting the definition of a rhetorical-scalable parameter that measures expressive
growth speed in bits per stage. A direct entropy measure shows that hierarchical selection over smaller
subsets markedly reduces choice uncertainty compared with flat selection across all modes. These
considerations appear to  transform static and non-measurable rhetorical taxonomies into more dynamic
and more measurable systems for discourse design. From this work, it would be possible to identify
a pathway for future AI systems to operate not only on language tokens but on layered rhetorical
reasoning structures, bridging linguistic, pedagogical, academic, and computational research.
}

\section{Introduction}

\label{sec:intro}

Rhetorical modes---also known as patterns of development---have long served as
fundamental templates for organizing discourse. From Aristotle's \emph{On
Rhetoric} (2007) to the modern pedagogical tradition established by Bain
(1866) and Hill (1895), rhetorical instruction evolved from oratorical
training toward written composition organized by recognizable textual forms.
Connors (1997) provides the most detailed account of this transformation,
showing how nineteenth- and twentieth-century composition pedagogy shifted
from persuasive appeals (ethos, pathos, logos) to structured \textquotedblleft
modes of discourse.\textquotedblright\

Several major composition textbooks (e.g., Corbett \& Connors 1999; Kirszner
\& Mandell 1986; Nadell, Langan \& Coxwell-Teague; Lunsford 2015, 2021; Hacker
\& Sommers 2020) promote an overlapping cluster of rhetorical
modes---narration, description, exemplification-illustration, definition,
classification-division, comparison-contrast, cause-effect, process analysis,
argument-persuasion, and exposition. Subsequent works (e.g., Oshima \& Hogue
2007; Smalley, Ruetten \& Kozyrev 2011) expand this cluster to include
additional modes such as analysis, synthesis, evaluation, and problem--solution.

\begin{figure}[ptb]
\centering\includegraphics[width=0.7\textwidth]{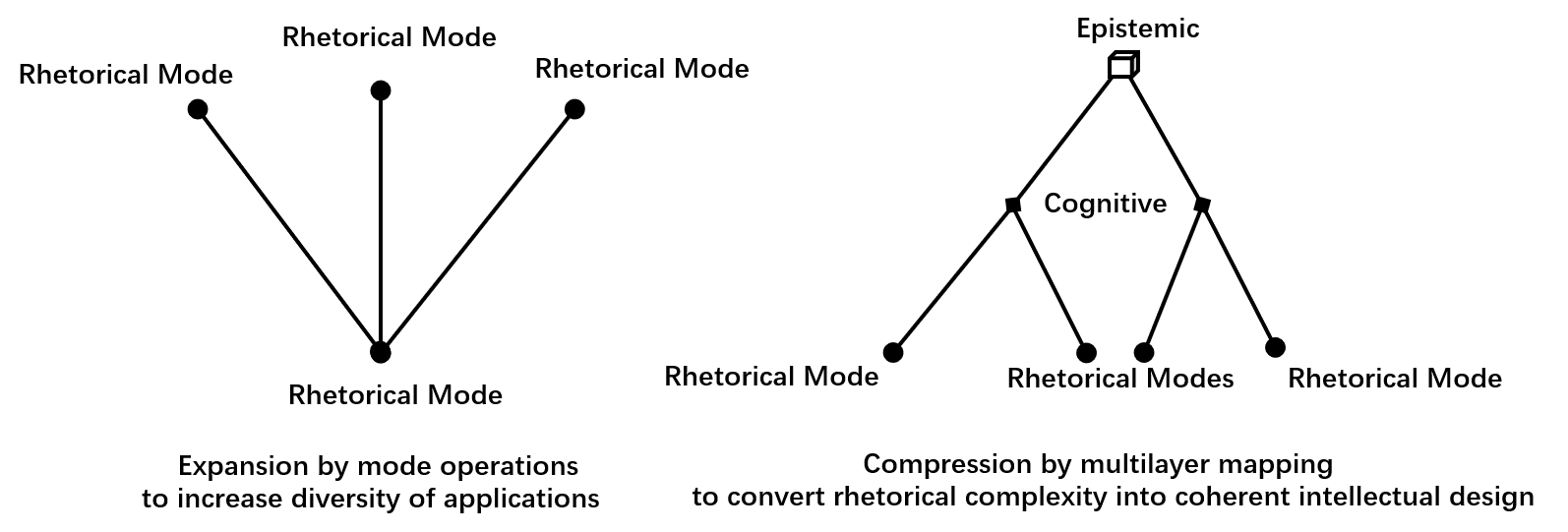}\caption{Expected
pathway in rhetorical mode development. Left: the growth and recombination of
modes from existing families to increase epistemic diversity. Right: the
mapping of rhetorical modes to epistemic purposes via an intermediate
cognitive layer.}%
\label{fig:ec}%
\end{figure}

Because rhetorical modes retain pedagogical importance (for example, they
continue to underpin college-level writing instruction), one might reasonably
expect that scholars would have examined them as interrelated cognitive
operations and, ultimately, as contributing to epistemic ends. Figure
\ref{fig:ec} sketches an expected developmental pathway. Because rhetorical
modes remain central to pedagogy, scholars have long called for
theory-building that links inventional heuristics to composing processes
(Young, 1978; Lauer, 2004; Bawarshi, 2003). Cognitive-process and
metacognitive research shows composing as recursive planning, translating, and
monitoring (Flower \& Hayes, 1981; Flavell, 1979), while discourse- and
genre-based work emphasizes purpose, audience, and contextual framing
(Bazerman, 1988; Beaufort, 2007; Hoey, 1983). At the same time, quantitative
and information-theoretic traditions (Shannon, 1948; Zipf, 1949; Lotman, 1990;
Simonton, 2004) reveal exponential/combinatorial patterns in language and
cognition, but, according to the knowledge of the present author, these
insights have not yet been widely extended to model rhetorical structure;
together these literatures suggest a promising basis for an integrated,
measurable account of modes $\rightarrow$ cognition $\rightarrow$ epistemic ends.

Rhetorical modes are recurrent, rule-governed patterns of discourse---stable
configurations of moves, constraints, and affordances that writers deploy to
achieve communicative ends (McQuade \& Atwan, 1998). Because a mode is a
pattern rather than a unitary string, it has formal properties (shape:
sequence of moves; size: scope; orientation: foregrounding vs. backgrounding;
extension: span across a text; edges: limits), which make it amenable to
systematic operations such as splitting, combining, reversing,
contracting/expanding, or bifurcating. Crucially, modes are typically oriented
toward epistemic ends (e.g., to communicate, to evaluate, to teach, to
discover), and those ends show patterned regularities; therefore modes can be
mapped to epistemic purposes. Introducing an intermediate functional layer
between surface form and epistemic goal reduces representational complexity
and makes a three-layer architecture (rhetorical modes $\rightarrow$ cognitive
or functional layer $\rightarrow$ epistemic purpose) both tractable and
pedagogically useful.

In Section 2 we explore extending the set of base rhetorical modes by
generating new modes through mode operations (e.g., split-unite,
forward-backward), motivated by demands from academic writing, learning
analytics, AI-assisted discourse modeling, and multimodal communication.
Building on a canonical set of fourteen modes, we propose duality operations:
many modes are inherently diatomic (e.g., comparison--contrast,
classification--division) and can be decomposed into atomic components or
recomposed into new hybrids; other modes can be paired with functional
counterparts to form a more symmetrical repertoire. We formalize diversity and
combination growth with simple combinatorics (binomial coefficients) and
introduce a tentative rhetorical-scalable parameter that captures combinative
capacity and measurable growth rates. A pedagogical \textquotedblleft
rhetorical cone\textquotedblright\ links this expansion to educational stages
(from preschool to graduate study), showing how repertoire grows with instruction.

Because unconstrained expansion risks producing non-linguistic or cognitively
unstable operations, Section 3 develops a three-layer cognitive mapping in
which surface modes project through an intermediate functional layer
(cognitive function layer) to epistemic purposes; this mapping constrains
redundancy, reduces cognitive entropy, and preserves epistemic coherence.
Together, these elements provide a tractable, operational framework with
possible pedagogical and computational applications.

Table \ref{tab:base_modes} lists the commonly seen rhetorical modes (in
alphabetical order), that will be considered as the base rhetorical modes for
mode operations and multilayer mapping.

\begin{table}[ptb]
\caption{Concise Summary of Base Rhetorical Modes (Rm)}%
\label{tab:base_modes}
\centering
{\small \setlength{\tabcolsep}{4pt} \renewcommand{\arraystretch}{1.2}
\begin{tabular}
[c]{@{}p{1.0cm}p{3.2cm}p{6.2cm}}%
\toprule \textbf{Code} & \textbf{Name} & \textbf{Core Function / Scope}\\
\midrule Rm(1) & Analysis--Synthesis & Decomposing concepts or systems and
integrating parts into coherent understanding.\\
Rm(2) & Analogy & Explaining ideas through structural or functional
similarity.\\
Rm(3) & Argument--Persuasion & Reasoning to demonstrate validity and influence
belief or action.\\
Rm(4) & Cause--Effect & Linking antecedent factors with resulting outcomes or
consequences.\\
Rm(5) & Classification--Division & Grouping entities by shared traits or
partitioning wholes into component parts.\\
Rm(6) & Comparison--Contrast & Analyzing similarities and differences among
entities or ideas.\\
Rm(7) & Definition & Specifying the meaning or essential boundaries of a
concept.\\
Rm(8) & Description & Depicting observable or sensory features to create vivid
representation.\\
Rm(9) & Evaluation & Applying explicit criteria to assess quality, value, or
effectiveness.\\
Rm(10) & Exemplification--Illustration & Supporting general claims with
specific instances or evidence.\\
Rm(11) & Exposition & Systematically clarifying information or ideas for
transparency.\\
Rm(12) & Narration & Presenting events or experiences in temporal or logical
sequence.\\
Rm(13) & Problem--Solution & Identifying an issue and proposing or justifying
a means of resolution.\\
Rm(14) & Process Analysis & Explaining ordered steps or stages that accomplish
a task or outcome.\\
\bottomrule &  &
\end{tabular}
}\end{table}

\section{Rhetorical Mode Expansion through Duality-based Mode Operation}

The extension of the base rhetorical modes is realized through properly
selected duality-based mode operations. The key point for mode operation is to
find the duality between two rhetorical modes (of which one is old, one is
new, or both are old).

\subsection{Duality-based Mode Operations}

\subsubsection{Introduction of mode operations}

Let $\mathcal{R}_{m}=\{R_{m}(1),R_{m}(2),\dots,R_{m}(K_{0})\}$ denote the set
of base or canonical rhetorical modes (e.g., narration, description,
definition, classification-division shown in Table \ref{tab:base_modes}, where
$K_{0}=14$), which can be divided into $7$ atomic rhetorical modes
($R_{m}^{(a)}(k)$) (like narattive, description, definition) and $7$ diatomic
rhetorical modes or compound rhetorical modes ($R_{m}^{(c)}(k)$) (like
classification-division, cause-effect, exemplification-illustration,
argumentation-persuation, problem-solution, comparision-constrast).

\begin{figure}[ptbh]
\centering
\includegraphics[width=0.8\textwidth]{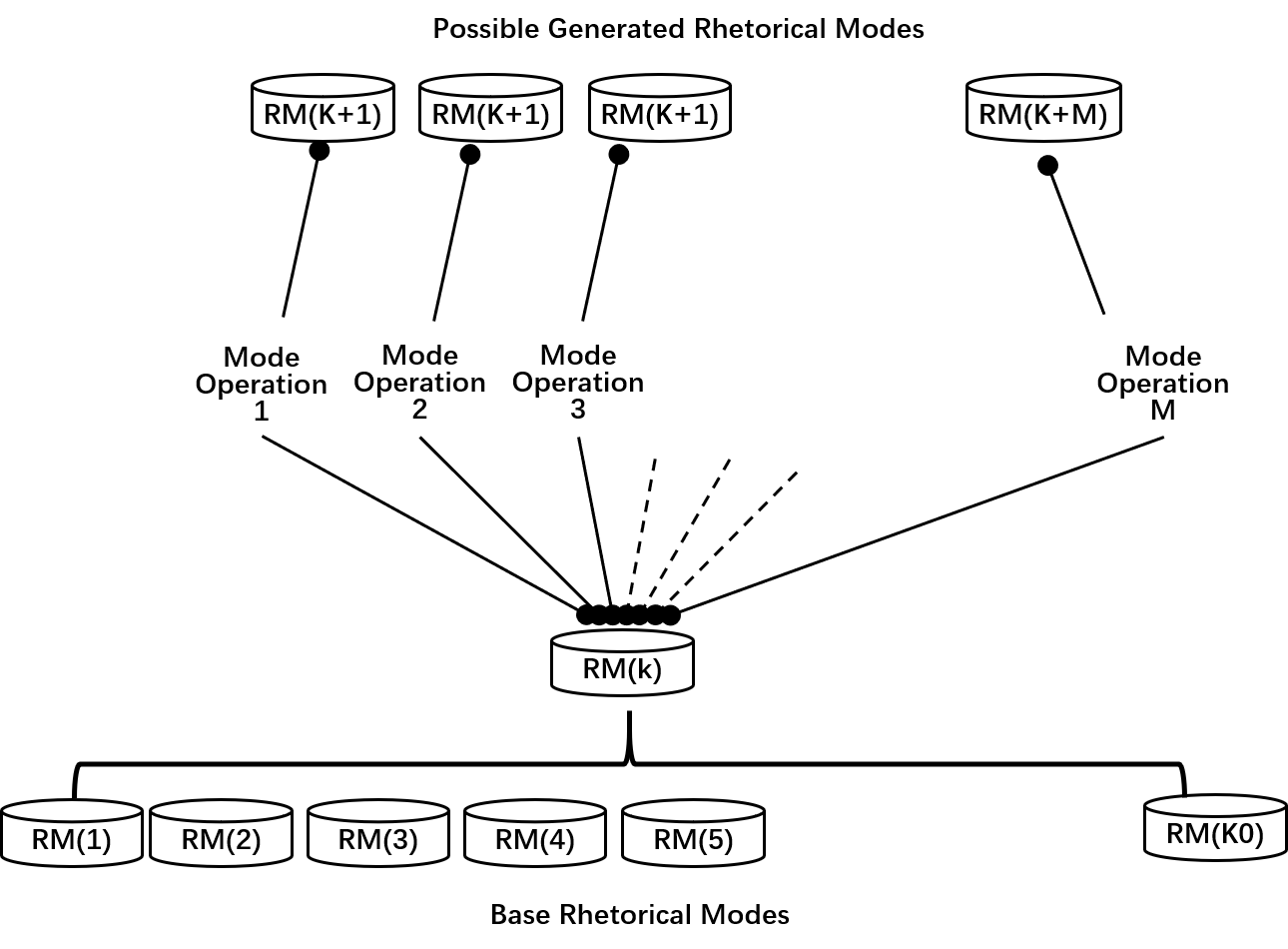}\caption{Schematic display of
rhetorical mode operations, generating generated rhetorical modes from base
rhetorical modes }%
\label{fig:mo}%
\end{figure}

Figure \ref{fig:mo} is an illustration of mode generation from operations done
on base rhetorical modes. The modes thus built are called generated modes or
extended modes, which may become rhetorical modes to be added to the existing
list of rhetorical modes (Table \ref{tab:base_modes} ), or remain as primitive
congnitive modes that complement the functional spreading of rhetorical modes
in its proper sence. 

Below we introduce four mode operations, based on split-unite duality,
forward-backward duality, expansion-contraction duality and orthogonal
duality. These four ones are not exclusive, but in this paper we do not
consider other possible mode operations.

\subsubsection{Split-Unite Duality}

\begin{figure}[ptbh]
\centering\includegraphics[width=0.4\textwidth]{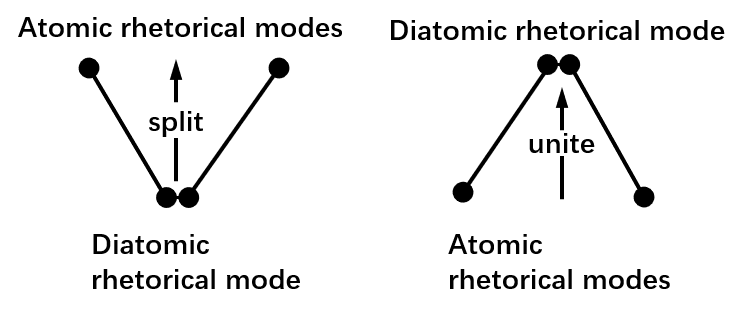}\caption{Schematic
display of split-unite operation}%
\label{fig:split-unit}%
\end{figure}

The split-unite duality suggests a split operation $\mathcal{O}%
_{\mathrm{split}}$ which splits a\ diatomic\ $R_{m}^{(c)}(k)$ into two atomic
rhetorical modes ($R_{m}^{(a)}$):
\[
\mathcal{O}_{\mathrm{split}}:\;R_{m}^{(c)}(k)\mapsto\{R_{m}^{(a)}%
(i),R_{m}^{(a)}(j)\}\text{,}%
\]
as illustrated in Figure \ref{fig:split-unit}. In Figure \ref{fig:split-unit}
we also illustrates unite operation ($\mathcal{O}_{\mathrm{unite}}$):
\[
\mathcal{O}_{\mathrm{unite}}:\;\{R_{m}^{(a)}(i),R_{m}^{(a)}(j)\}\mapsto
R_{m}^{(c)}(k)\}
\]
which produces a diatomic rhetorical mode from any two atomic rhetorical modes.

The application of $\mathcal{O}_{\mathrm{split}}$ to all $R_{m}^{(c)}(k)$ in
Table \ref{tab:base_modes} leads to 14 novel atomic rhetorical modes, as shown
in Table \ref{tab:decomposed_modes}. For instance, apply $\mathcal{O}%
_{\mathrm{split}}$ to \textbf{Classification-Division} leads to
\textbf{Classification} and \textbf{Dvision}. \begin{table}[ptb]
\caption{Concise Summary of Decomposed Rhetorical Modes (from Compound Modes)
}%
\label{tab:decomposed_modes}
\centering
{\small \setlength{\tabcolsep}{4pt} \renewcommand{\arraystretch}{1.2}
\begin{tabular}
[c]{@{}p{1.0cm}p{3.2cm}p{6.2cm}}%
\toprule \textbf{Code} & \textbf{Name} & \textbf{Core Function / Scope}\\
\midrule Rm(15) & Classification & Grouping entities or ideas by shared traits
or organizing principles.\\
Rm(16) & Division & Partitioning a whole into constituent parts to clarify
structure or function.\\
Rm(17) & Cause & Identifying antecedent factors or conditions producing a
specific result.\\
Rm(18) & Effect & Describing outcomes or consequences emerging from identified
causes.\\
Rm(19) & Exemplification & Supporting general claims with concrete examples or
representative cases.\\
Rm(20) & Illustration (Evidence) & Providing factual proof or documentation
that substantiates a claim.\\
Rm(21) & Argument & Advancing a logically supported claim based on reasons and
evidence.\\
Rm(22) & Persuasion & Influencing attitudes or actions through emotional,
ethical, or value-based appeal.\\
Rm(23) & Problem & Identifying a question, issue, or gap requiring
interpretation or resolution.\\
Rm(24) & Solution & Proposing and justifying a method or means to resolve an
identified problem.\\
Rm(25) & Comparison & Highlighting similarities among multiple entities,
ideas, or phenomena.\\
Rm(26) & Contrast & Highlighting and interpreting key differences among
entities, ideas, or phenomena.\\
Rm(27) & Analysis & Breaking a concept or system into parts to reveal its
logic or internal relationships.\\
Rm(28) & Synthesis & Integrating diverse perspectives or arguments into a
unified understanding.\\
\bottomrule &  &
\end{tabular}
}\end{table}

\begin{figure}[ptbh]
\centering\includegraphics[width=0.2\textwidth]{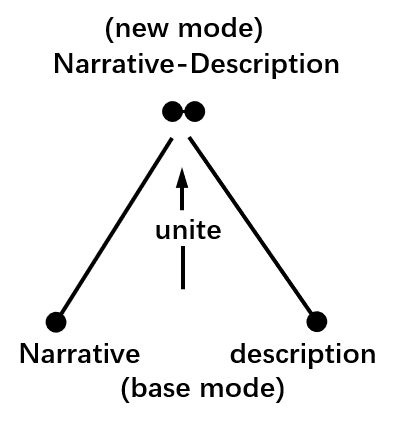}\caption{Schematic
display of model operation, using split-unite duality, leading to new mode
"narrative-description".}%
\label{basetonew1}%
\end{figure}

One may also apply the unite operation $\mathcal{O}_{\mathrm{unite}}$ to the
atomic rhetorical modes ($R_{m}^{(a)}(k)$) in Table \ref{tab:base_modes} and
Table \ref{tab:decomposed_modes}, producing for instance new compound mode
like "\textbf{narrative}-\textbf{description}", see Figure \ref{basetonew1}.

The \textbf{narration--description} mode is generated through the Unite
operation of the Split--Unite Duality, combining the atomic modes
\textbf{narration} and \textbf{description} into a single diatomic form. While
narration organizes events in temporal sequence and description depicts the
concrete qualities of each event or scene, their unification produces a
discourse structure in which actions unfold over time while each stage is
vividly rendered in detail. This mode enables readers to perceive both process
and texture---to follow a sequence while simultaneously visualizing its
internal states. Cognitively, it integrates sequential reasoning with
perceptual grounding, facilitating stronger memory encoding and conceptual
understanding. In practical application, the narration--description mode is
particularly effective in teaching demonstrations, experimental reporting, and
case-based explanations, where dynamic processes must be conveyed with both
chronological clarity and descriptive precision.

If all choices are possible among the 7 atomic rhetorical modes in Table
\ref{tab:base_modes}, then we would have
\[
\binom{7}{2}=\frac{7!}{2!(7-2)!}=21
\]
new possibilities for diatomic\ $R_{m}^{(c)}(k)$.

If further, we consider all the atomic rhetorical modes, including the seven
ones in Table \ref{tab:base_modes} and the 14 novel atomic rhetorical modes,
as shown in Table \ref{tab:decomposed_modes}, then we have in total $21$
atomic rhetorical modes, and the possibiliy to have diatomic modes
($R_{m}^{(c)}$) by unite operation is%
\[
\binom{21}{2}=210
\]

A triatomic mode may be defined as the composition of three atomic modes. We
do not develop this case separately since it is a direct extension of the
diatomic generation rule. Section 3 formalizes academic functions that couple
one core mode with supplementary modes; triatomic compositions are therefore
covered by that formulation.

\subsubsection{Forward-Backward Duality}

This operation reverses the temporal or logical direction of a mode:
\[
\mathcal{F}_{R}(R_{m}(k))=R_{m}^{-1}(k)
\]
For instance, apply this duality operation generates from \textbf{Cause} to
\textbf{Effect}, $\quad$\textbf{Problem} to \textbf{Solution}, and vice versa.
These are still among the atomic rheotrical modes presented in Table
\ref{tab:decomposed_modes}.

\begin{figure}[ptbh]
\centering\includegraphics[width=0.55\textwidth]{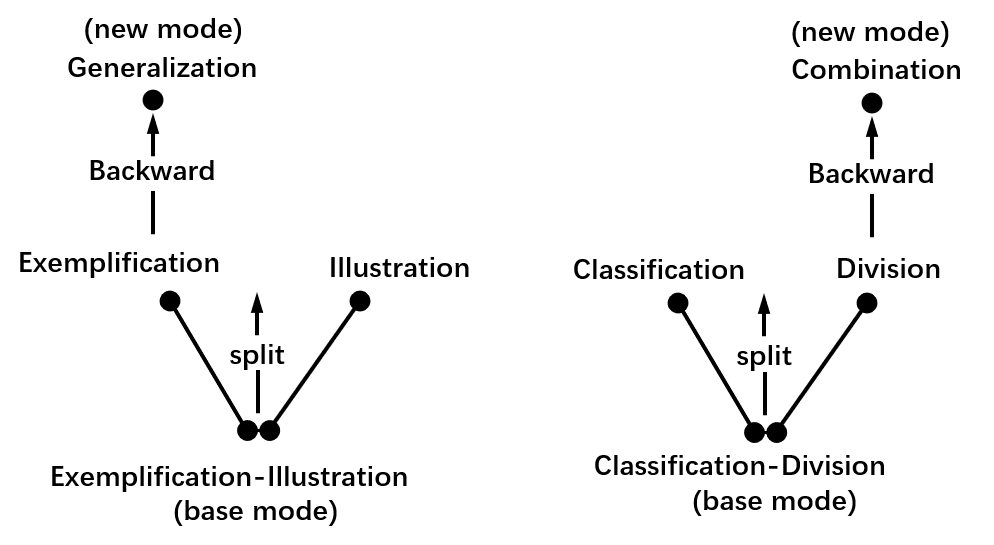}\caption{Schematic
display of model operation, using split-unite duality and forward-backward
duality, leading to generated mode \textbf{generalization} and
\textbf{combination}.}%
\label{fig:basetonew2}%
\end{figure}

However, if this operation is applied to \textbf{Exemplification}, we would
have \textbf{Generalization}, which seems to be a new rhetorical mode. In
fact, a special case for Forward-Backward Duality is Deduction-induction
Duality, which means reasoning from \textbf{Generalization} to
\textbf{Exemplification} (deduction) and vice versa. Furthermore, if this
operation is applied to \textbf{Division}, we would have \textbf{Combination}
as a generated mode. See Figure \ref{fig:basetonew2} for this kind of
operation leading to generated modes.

We do not discuss whether this operation generates more generated modes if
applied to other atomic modes (the reader may use this operation to do more).

\subsubsection{Expansion-Reduction Duality}

This operator adjusts the scale or granularity of a mode:
\[
\mathcal{E}_{\mathrm{expand}}(R_{i})=R_{i}^{+},\qquad\mathcal{E}%
_{\mathrm{reduce}}(R_{i})=R_{i}^{-},
\]
where $R_{i}^{+}$ provides elaboration or amplification, while $R_{i}^{-}$
compresses or summarizes the discourse. For example, \textbf{Exposition}
corresponds to expansion, by Expansion-Reduction Duality, we obtain from
\textbf{Exposition} the generated mode \textbf{Summary}.

\subsubsection{Orthogonal Duality ($\mathcal{O}_{\!\perp}$)}

Two modes are orthogonal if they operate in distinct cognitive dimensions,
e.g., horizontal (temporal) versus vertical (structural):
\[
\mathcal{O}_{\!\perp}(\text{Narration})=\text{Description},\quad
\mathcal{O}_{\!\perp}(\text{Description})=\text{Narration}.
\]
This duality models complementary perceptual axes sequence versus depth.

\subsubsection{Rhetorical completeness and the need to have operator calculus}

The four mode operations, while not exclusive, explain both the familiar
dualities of classical rhetorical modes and how generated modes can be derived
toward rhetorical completeness. Some of the generated modes coincide with
categories already used in rhetorical practice; others fall outside the
classical sense but nevertheless exhibit coherent patterns and useful
applications. Such departures from classical form do not preclude these modes
from having well-defined structure or practical value.

Together, the four and possible other mode operations could provide a
mechanism for discovering, pairing, and extending rhetorical modes. It is
hopeful that, by iterative use, mode operations may help to generate a
balanced and potentially open-ended rhetorical system, in which expressive and
cognitive capacities increase through structural symmetry and logical reversibility.

However, to achieve the above purpose more fomally, it would be useful to
develop, in the future, operator calculus, like in comptional modelling, that
can abstract the logical and functional relations among rhetorical modes. More
precisely, the full specification of four or even more mode operations---each
expressed as a symbolic operator with definable algebraic behavior---could be
assigned symbolic operators that are programmable for computer aided searching
of more rhetorical modes, and then put rhetorical modes as patterns of
language that help to increase the efficiency of large Language Models.

\subsection{Rhetorical Scalable\textit{\ }Parameter Measuring the Relative
Rhetorical Capacity}

Rhetorical modes are not merely names in a textbook; they are patterned,
reusable operations for developing ideas in discourse. Treating modes as
measurable objects makes it possible to ask quantitative questions that are
pedagogically and scientifically useful: How many distinct rhetorical choices
does a text offer? How quickly do new modes increase the expressive
possibilities of a course or curriculum? At what pace does rhetorical growth
risk overwhelming learners? The rhetorical-scalable parameter discussed here
formalizes these questions.

\subsubsection{Rhetorical combination capacity and marginal rhetorical bit
(MRB)}

A rhetorical-scalable parameter measures the relative rhetorical growth speed,
so we need an inherent parameter that remains to be a constant.

The number $K$ of rhetorical modes ($\mathcal{R}_{m}$), available in a system
(an academic paper, a course, a genre), is now considered as a variable and
called the rhetorical width. This width represents the expressive
dimensionality or \textit{linguistic bandwidth} of the discourse universe. A
system with higher rhetorical width allows more expressive directions and
functional differentiation, analogous to degrees of freedom in mechanics or
dimensions in geometry.

At the level of an individual discourse unit (a sentence, paragraph, or
slide), one typically use only a small subset ($k$) of the $K$ modes. The
number of distinct ways to choose exactly $k$ modes from $\mathcal{R}_{m}$ is
the binomial coefficient%
\begin{equation}
K_{C}(k,K)=\binom{K}{k}=\frac{K!}{k!(K-k)!} \label{eq-0}%
\end{equation}
This quantity measures local expressive richness: small $k$ indicates
single-mode or narrowly focused discourse; larger $k$ indicates dense,
multimodal units. This counts how many unique combinations of modes can occur
at each structural level of discourse and measures the local expressive
richness or compositional order of a text. Small $k$ corresponds to near
single-mode writing; large $k$ represents complex, multimodal discourse.

The maximum number $K_{\max}(K)=K_{C}(k,K)$ of $k$-combinations occurs at
$k=k_{m}$ with%

\[
k_{m}=\left\lfloor \frac{K}{2}\right\rfloor \approx\frac{K}{2}%
\]
and this peak value ($K_{\max}(K)=K_{C}(k_{m},K)$) is a simple proxy for
rhetorical complexity. The parameter $K_{\max}(K)$ measures the maximum
combinational potentional since it corresponds to the greatest number of
distinct multimodal configurations that can coexist at equilibrium.

The total number of (non-empty) rhetorical combinations (NRC) that can be
formed from those $K$ modes is%
\begin{equation}
K_{NRC}(K)=\sum_{k=1}^{K}\binom{K}{k}=2^{K}-1. \label{eq01}%
\end{equation}
The quantity of $K_{NRC}(K)$ shows rhetorical capacity grows exponentially
with the number of modes: every added mode doubles the set of possible
combinations (approximately --- the \textquotedblleft$-1$\textquotedblright%
\ is negligible for moderate-to-large $K$). Because exponential growth is hard
to reason about intuitively, we work on the log scale. Using base-2
logarithms, we define the rhetorical capacity $K_{RC}(K)$ as%

\begin{equation}
K_{RC}(K):=\ln_{2}K_{NRC}(K)=K \label{eq-1}%
\end{equation}
so each additional mode contributes one bit of new combinatorial expressive
capacity. This gives an immediate, interpretable metric: adding one new mode
doubles the number of possible rhetorical configurations. Thus, $K_{RC}(K)$
quantifies the log-scale expressive potential of a system: one extra mode one
extra bit of rhetorical capacity.

Now we are ready to identify an inherent parameter that remains to be a
constant with variable $K$. This inherent parameter is the Marginal Rhetorical
Bit (MRB), defined as the derivative of $K_{RC}(K)$ with respect to $K$%
\[
\text{MRC:=}\frac{dK_{RC}(K)}{dK}%
\]
Because of (\ref{eq-1}) we have%
\begin{equation}
\text{MRC}=1\text{ (bit per mode)}\label{eq-2}%
\end{equation}
so this parameter is indeed a constant.

The parameter MRB measures how many bits of additional rhetorical capacity are
gained by adding one mode. Because MRB $\approx1$, each new mode contributes
roughly one bit (i.e., doubles the space of configurations).

\subsubsection{Rate of rhetorical introduction (RRI)}

Rhetorical modes can be introduced gradually through progress of education,
writing and reasoning. A stage of education, like elementary school, may
introduce two modes over six years; a writing course may introduce one or two
new rhetorical modes per week; a writing program may add three modes per year.

To quantify the progress, it is convient to introduce the rate of rhetorical
introduction (RRI), to be denoted by $L_{n}$.\ \ The rate of rhetorical
introduction ($L_{n}$) is an empirical, directly observable quantity. It
refers to the average number of new rhetorical modes introduced per stage
(stage = paragraph, hour, semester, year, etc. --- choose the unit that fits
your application)%
\[
L_{n}=\frac{\text{new rhetorical modes introduced}}{\text{stage}}\text{
(mode/stage)}%
\]

Table \ref{tab:mode_acquisition} shows possible values of $L_{n}$ during each
educational stage.

\begin{table}[ptbh]
\caption{Typical cumulative acquisition path of rhetorical modes across
educational levels}%
\label{tab:mode_acquisition}
\centering
{\small
\begin{tabular}
[c]{@{}p{3cm}p{1.2cm}p{5cm}p{2.9cm}}%
\toprule \textbf{Educational Stage} & $K$ & \textbf{New Modes Introduced} &
\textbf{$L_{n}$ (Mode/Year)}\\
\midrule \textbf{KG (Preschool)} & 1--2 & Narration, Description & $L_{n}<
0.33 $\\
\textbf{Elementary School} & 4--5 & Exemplification, Definition,
Classification & $L_{n}\approx0.33$\\
\textbf{Middle School} & 7--8 & Comparison, Contrast, Cause--Effect, Process,
Argument & $L_{n}\approx0.66$\\
\textbf{High School} & 10--12 & Analogy, Illustration, Evaluation,
Problem--Solution, Persuasion & $L_{n}\approx0.66$\\
\textbf{Undergraduate (Y1--Y4)} & 14--16 & Analysis, Synthesis, Exposition,
Division--Classification (dual use) & $L_{n}\approx1$\\
\textbf{Graduate (Master / PhD)} & $>20$ & All canonical + dual expansions
(Combination, Generalization, etc.) & $L_{n}>1$\\
\bottomrule &  &  &
\end{tabular}
}\end{table}

\subsubsection{Rhetorical-scalable parameter: RRI divided by MRB}

Define the rhetorical-scalable parameter $R_{scale}$ as $L_{n}$ (mode per
stage) multiplied by MRB (bit per mode),%
\[
R_{scale}:=L_{n}\times\text{MRB (bit per stage)}%
\]
Because of MRB=1 (bit per mode), it follows algebraically that%
\[
R_{scale}=L_{n}%
\]
Thus the rhetorical-scalable parameter equals the rate of rhetorical
introduction (RRI) and can be read directly as bits of rhetorical capacity
added per stage. The rhetorical-scalable parameter is therefore an immediate,
interpretable measure: if $R_{C}=0.5$, you add half a bit per stage (one new
mode every two stages); if $R_{C}=2$, you add two bits per stage (two new
modes every stage).

\subsubsection{Practical implications and normalization.}

Reporting $R_{scale}$ is operational: instructors or designers can prescribe
or estimate $L_{n}$ and so immediately quantify rhetorical growth in
bits/stage. Because MRB is constant, comparisons across curricula or
interventions are straightforward.

For cognitive-load diagnosis, normalize $R_{scale}$ against an empirically
determined processing capacity $C_{0}$ (bits/stage a learner can assimilate).
Define normalized load $R_{scale}^{\ast}=R_{sclae}/C_{0}$. The values
\[
R_{scale}^{\ast}>1
\]
indicate potential overload. \ Subcritical load means $R_{scale}^{\ast}<1$,
critical load means $R_{scale}^{\ast}=1$, and supercritical load means
$R_{scale}^{\ast}>1$.

\subsection{Rhetorical mode cone: scalable pedagogy that aligns with levels of
cognitive development}

Table \ref{tab:rhetcomplex} gives the maximum number of combination $K_{\max
}(K)$ (and the corresponding $k_{m}$) and the total number of rhetorical
combinations\ ($K_{NRC}$) for rhetorical mode width $K$ up to 30.

\begin{table}[ptbh]
\caption{Rhetorical parameters ($k_{m},K_{\max},K_{NRC}$) for rhetorical width
up to $K=1$--$30$}%
\label{tab:rhetcomplex}%
\centering
{\small
\begin{tabular}
[c]{ccrr|ccrr}\hline
$K$ & $k_{m}$ & $K_{max}(K)$ & $K_{NRC}(K)$ & $K$ & $k_{m}$ & $K_{max}(K)$ &
$K_{NRC}(K)$\\\hline
1 & 0 & 1 & 1 & 16 & 8 & 12,870 & 65,535\\
2 & 1 & 2 & 3 & 17 & 8 & 24,310 & 131,071\\
3 & 1 & 3 & 7 & 18 & 9 & 48,620 & 262,143\\
4 & 2 & 6 & 15 & 19 & 9 & 92,378 & 524,287\\
5 & 2 & 10 & 31 & 20 & 10 & 184,756 & 1,048,575\\
6 & 3 & 20 & 63 & 21 & 10 & 352,716 & 2,097,151\\
7 & 3 & 35 & 127 & 22 & 11 & 705,432 & 4,194,303\\
8 & 4 & 70 & 255 & 23 & 11 & 1,352,078 & 8,388,607\\
9 & 4 & 126 & 511 & 24 & 12 & 2,704,156 & 16,777,215\\
10 & 5 & 252 & 1,023 & 25 & 12 & 5,200,300 & 33,554,431\\
11 & 5 & 462 & 2,047 & 26 & 13 & 10,400,600 & 67,108,863\\
12 & 6 & 924 & 4,095 & 27 & 13 & 20,058,300 & 134,217,727\\
13 & 6 & 1,716 & 8,191 & 28 & 14 & 40,116,600 & 268,435,455\\
14 & 7 & 3,432 & 16,383 & 29 & 14 & 77,558,760 & 536,870,911\\
15 & 7 & 6,435 & 32,767 & 30 & 15 & 155,117,520 & 1,073,741,823\\\hline
\end{tabular}
}\end{table}

For $K=10$, we have $k_{m}=\allowbreak5$, $K_{\max}=\allowbreak252$ and
$K_{NRC}=2^{10}-1=1{,}023$. If the rhetorical width is increased to $K=16$,
then $k_{m}=8$, $K_{\max}=\allowbreak12,870$ and $K_{NRC}=65{,}535$. This
gives a ratio $\frac{K_{NRC}(16)}{K_{NRC}(10)}\approx64$, showing the role of
adding the extensions.

If the rhetorical width is further increased to $K=20$, then
\[
k_{m}=10,\text{ }K_{\max}=\allowbreak184,756,\text{ }K_{NRC}=1{,}048{,}575.
\]
This gives a ratio $\frac{K_{NRC}(20)}{K_{NRC}(10)}\approx1{,}025$. $\ $

If we want $K_{NRC}$ to be just more than $10$ millions, then $K>\frac
{\ln(10^{8}+1)}{\ln2}\simeq\allowbreak27$.

Thus, as $K$ grows, the number of possible non-empty configurations rises
exponentially---from 1,023 at $K=10$ to 1,048,575 at $K=20$ to $1.34\times
10^{8}$ at $K=27$. This formalizes the intuition that adding modes does not
merely add options linearly; it \textit{multiplies} the expressive space.
Thus, the total cognitive intensity grows exponentially with the rhetorical
width $K$, each additional mode disproportionately expands the expressive
potential of discourse. This quantifies how extending and balancing the system
of modes greatly increases the functional capacity of language.

The the proportion of available rhetorical modes actually employed in a text
could be\ evaluated through identifying the Rhetorical Mode Coverage%
\[
C_{m}=\frac{K_{u}}{K}\text{.}%
\]

When $C_{m}\rightarrow1$, the text demonstrates high rhetorical diversity,
integrating nearly all available modes (typical of advanced academic or
pedagogical writing). When $C_{m}\approx0.5$, the text has moderate rhetorical
variety, common in undergraduate or professional prose. When $C_{m}<0.3$, the
text has limited rhetorical deployment, typical of early educational writing
or narrowly technical genres. Rhetorical Mode Coverage provides a direct,
quantifiable index of rhetorical maturity across developmental stages. For
instance, a middle-school essay employing narration, description, and
exemplification may have $C_{m}=0.15$, whereas a graduate thesis integrating
definition, classification, comparison, cause--effect, evaluation, and
synthesis may reach $C_{m}=0.8$. Thus, $C_{m}$ allows teachers and researchers
to map rhetorical development longitudinally across education levels.

Thus $K,K_{\max}$ and $K_{NRC}$ offer a \textit{scalable pedagogy} that aligns
with levels of cognitive development.

\begin{figure}[ptbh]
\centering
\includegraphics[width=0.8\textwidth]{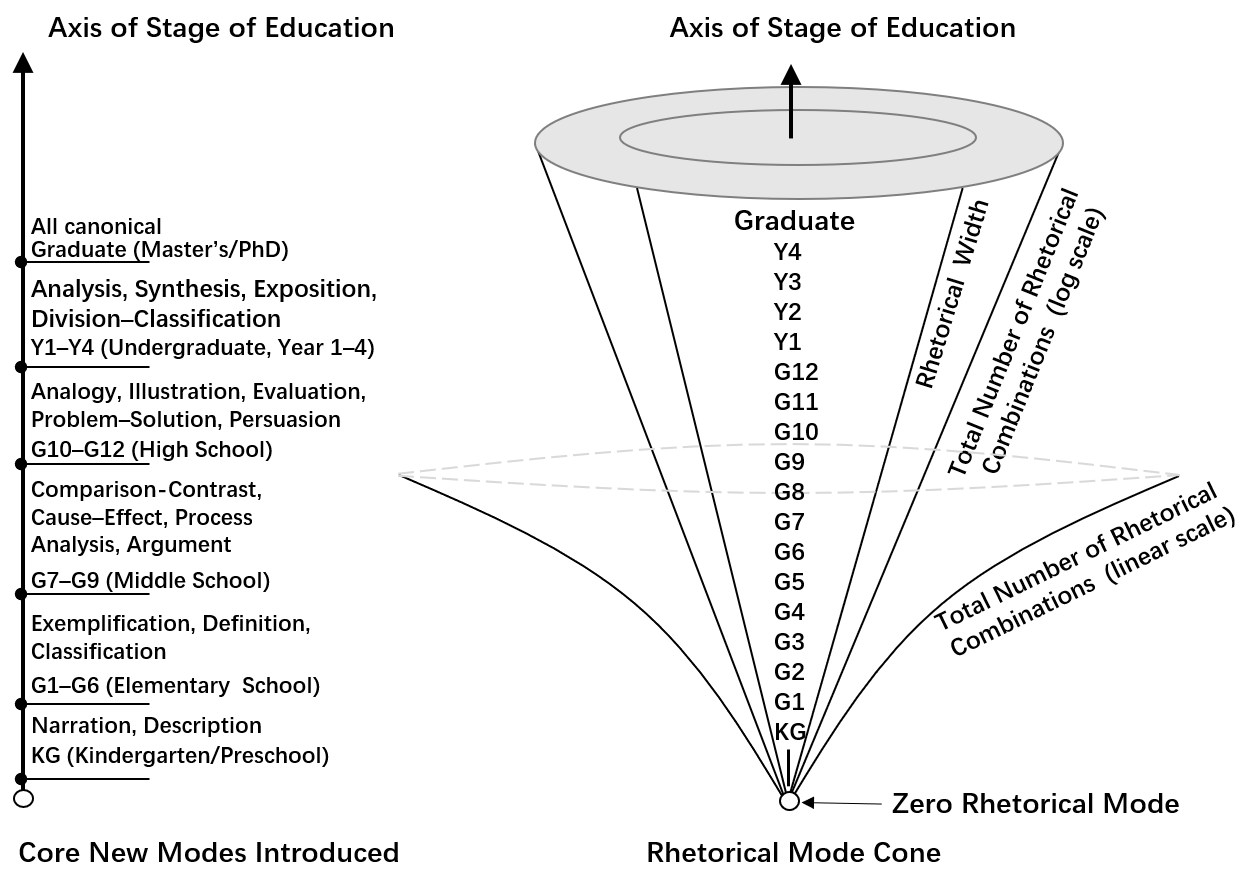}\caption{Rhetorical Mode Cone,
displaying educational related growth in Rhetorical Width ($K$) and the number
of (non-empty) rhetorical combinations ($K_{NRC}$) (log compressed and
unlogged). }%
\label{fig:rmcone}%
\end{figure}

Figure \ref{fig:rmcone} defines a rhetorical mode cone, showing how rhetorical
widths ($K$) and other combinative parameters increase with education stage.

%In your preamble:
%\usepackage{tikz}
%\usetikzlibrary{calc}

Here we have provided some quantities using binomial coefficient to show how
the expressive potential increases exponentially with the number of modes,
demonstrating that the richer the system, the greater the cognitive and
expressive space it opens. Yet, such expansion also implies higher
entropy---that is, greater uncertainty and cognitive demand when selecting or
interpreting rhetorical forms.

\section{Pyramid Multilayer Mapping and Academic Applications}

To counterbalance the growth of complexity with extended rhetorical modes, we
attempt to introduce the concept of multilayer mapping, showing that
structured mappings between rhetorical, cognitive, and epistemic layers can
reduce entropy---the informational uncertainty that burdens both writers and
readers. Since each element of an upper layer corresponds to several elements
of an adjacent lower layer. the shape is like a pyramid, so we call this
Pyramid Shape Multilayer Mapping (Pyramid Multilayer Mapping for short).

\subsection{Multilayer Mapping}

The expansion of rhetorical modes through duality operations presented in
Section 2 reveals a vast expressive landscape whose combinatorial richness
grows exponentially. Yet this same richness also multiplies the number of
possible rhetorical configurations, producing higher cognitive and
informational entropy (uncertainty). Without an organizing mechanism, writers
and readers must navigate an almost unbounded field of choices, increasing the
mental effort required to select, structure, and interpret discourse.

To address this challenge, an efficient way is to introduce a pyramid
multilayer mapping framework that connects rhetorical modes to final
applications. Here we consider such a framework that connects three layers:
rhetorical, cognitive, and epistemic layers.

It is knowen from information-theoretic principles, that hierarchical mapping
acts as an entropy-reduction process. Here, such a mapping constrains
rhetorical variability by aligning linguistic operations with reasoning
functions and knowledge purposes. In this view, the act of academic writing
becomes a progressive transformation from disorder to order---from expressive
diversity to purposeful structure.

\begin{figure}[ptbh]
\centering\includegraphics[width=0.9\textwidth]{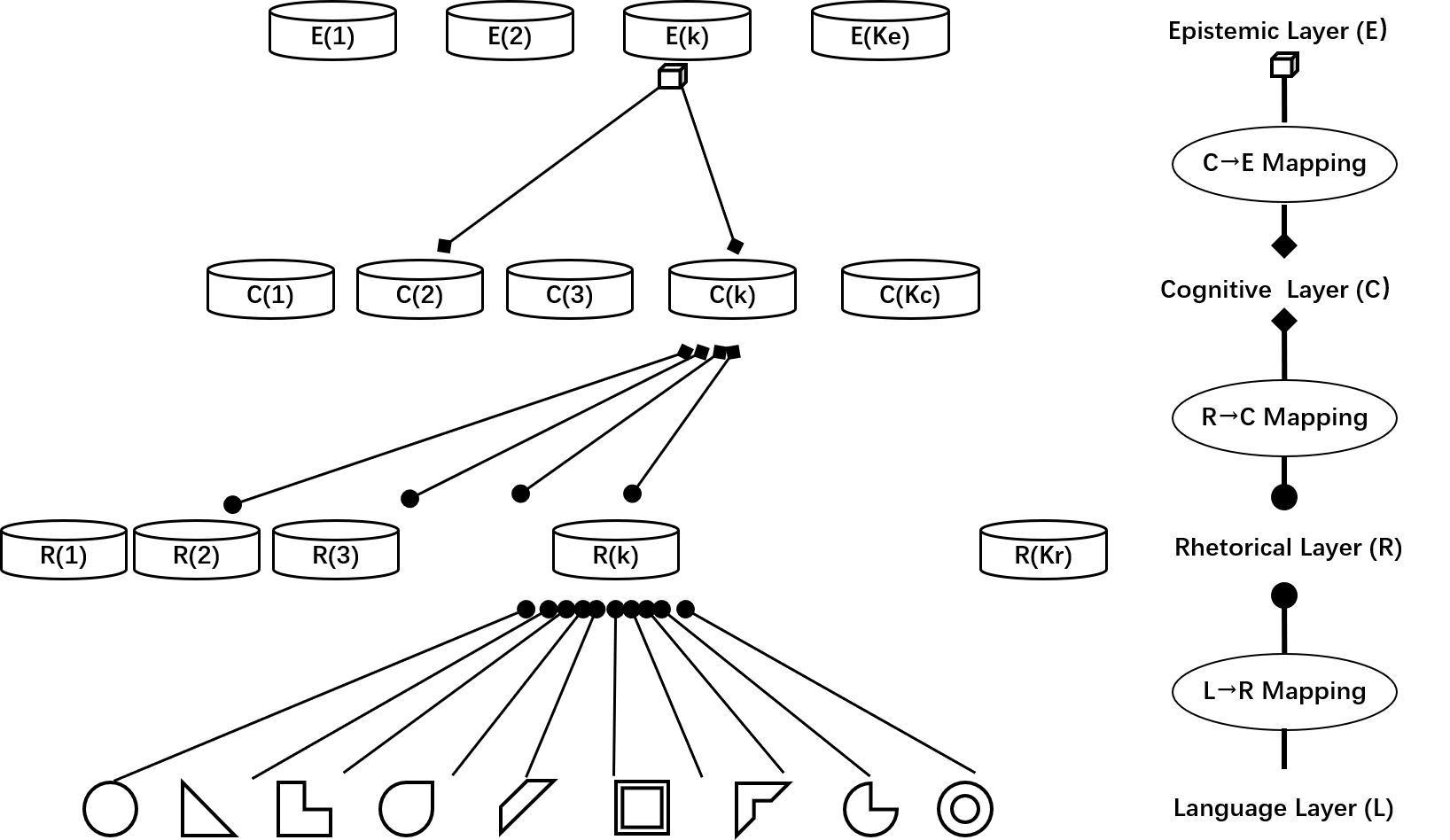}\caption{Schematic
display of three layer mapping in general pyramid multilayer mapping}%
\label{fig:rce}%
\end{figure}

Figure \ref{fig:rce} illustrates a three-layer mapping framework, mapping from
rhetorical modes (R) to cognitive functions (C), and laterly to epistemic
purposes (E),.

Formally, let $R=\left\{  r(1),r(2),\cdots,r(K_{R})\right\}  $, $C=\left\{
c(1),c(2),\cdots,c(K_{C})\right\}  $, and $E=\left\{  e(1),e(2),\cdots
,e(K_{E})\right\}  $. The three-layer mapping is realized through
$R\rightarrow C$, $C\rightarrow E$, and then $R\rightarrow E$.

Mapping between rhetorical, cognitive, and epistemic layers reduces cognitive
burden, and this reduction can be quantified through information entropy, if
mapping can be specified. Entropy provides a natural measure of
\textit{uncertainty} or \textit{choice space}---that is, how many possible
configurations a writer or reader must navigate.

Consider a rhetorical system with $K$ available modes. If each mode occurs
with probability $p_{i}$, the entropy of rhetorical choice is
\[
H_{R}(K) = -\sum_{i=1}^{K} p_{i} \log_{2} p_{i} .
\]
When all modes are equally likely, this simplifies to $H_{R} = \log_{2} K$.
For instance, with 18 canonical modes, we have $H_{R} \approx4.17$ bits,
indicating that a writer effectively chooses among about $2^{4.17} \!\approx\!
18$ rhetorical alternatives per unit of discourse.

\vspace{6pt} \noindent If we instead consider all possible subsets of these
modes, the number of choices of size $k$ is given by (\ref{eq-0}), and the
total number of nonempty subsets is given by (\ref{eq01}). The corresponding
probability mass function is therefore $p_{k}=\frac{K_{C}(k,K)}{K_{NRC}(K)}$,
and the entropy becomes
\begin{equation}
H_{R}(K)=-\sum_{k=1}^{K}\frac{\binom{K}{k}}{2^{K}-1}\log_{2}\!\left(
\frac{\binom{K}{k}}{2^{K}-1}\right)  . \label{eq-5}%
\end{equation}
This expression represents the Shannon entropy (in bits) of the distribution
of subset sizes when each nonempty subset of $K$ elements is equally probable.

\vspace{6pt}According to Cover \& Thomas (2006) and Knessl (1998), the
right-hand-side of (\ref{eq01}) has an asymptotic value $\frac{1}{2}\log
_{2}\!\left(  \frac{\pi eK}{2}\right)  $,
\[
H_{R}(K)\approx\frac{1}{2}\log_{2}\!\left(  \frac{\pi eK}{2}\right)  +o(1).
\]
Numerically, for $K=2$, $H_{R}\!\approx\!0.92$ bits; for $K=7$, $H_{R}%
\!\approx\!2.39$ bits; for $K=10$, $H_{R}\!\approx\!2.70$ bits; for $K=14$,
$H_{R}\!\approx\!2.95$ bits; for $K=18$, $H_{R}\!\approx\!3.13$ bits; for
$K=20$, $H_{R}\!\approx\!3.21$ bits; and for $K=100$, $H_{R}\!\approx\!4.37$ bits.

\noindent The reduction of entropy with smaller $K$ implies that introducing a
multilayer mapping here each layer involves a smaller number of choices, which
reduces the overall uncertainty of rhetorical selection.

For example, if the rhetorical-mode layer has $K=20$, then $H_{R}%
(20)\!\approx\!3.21$ bits. At the next layer, each cognitive function selects
among at most four rhetorical modes, giving $H_{R}(4)\!\approx\!1.81$ bits.
Similarly, each epistemic function selects from at most four cognitive
functions, again yielding $H_{R}(4)\!\approx\!1.81$ bits. The combined entropy
of these two layers is therefore considerably lower than a direct single-layer
selection from all $K=20$ modes, demonstrating the efficiency of hierarchical
or layered rhetorical organization. These numbers demonstrate why multilayer
mapping is valuable, though real entropy reduction depends also how we
organize the connection between different layers.

Rhetorical diversity expands expressive potential but also inflates cognitive
entropy. Cognitive and epistemic mappings act as structured constraints that
transform this diversity into intelligible order. The quantitative reduction
in entropy thus represents not information loss but \textit{knowledge
organization}---a measurable indicator of how academic discourse converts
linguistic variety into purposeful understanding.

\subsection{The Three Layer Mapping and Applications}

\subsubsection{Rhetorical Layer to Cognitive Layer}

The cognitive layer is supposed to contain $K_{C}$ cognitive functions,
denoted $C\left(  k\right)  ,k=1,2,\cdots,K_{C}$. It could be imagined that
\textbf{observe} is one such function, which means to register phenomena and
qualitative data, with the help of rhetorical modes (like description,
narration, exemplification and evidence) in the R layer. Table
\ref{tab:cognitive_mapping} listed $K_{C}=14$ tyical cognitive functions,
including the code, the function name, meaning or operation, and typical
rhetorical modes in mapping, i.e., typical rhetorical modes realizing the
cognitive functions. The $14$ cognitive functions listed in Table
\ref{tab:cognitive_mapping} are not exclusive. For some specific applications
we may have different functions as discussed in section 3.2.4, where we also
discuss how rhetorical modes can be refreshed to have more academic oriented
meanings. \begin{table}[ptbh]
\caption{Mapping between Cognitive Functions (C-layer) and Typical Rhetorical
Modes (R-layer)}%
\label{tab:cognitive_mapping}
\centering
\begin{tabular}
[c]{@{}p{0.9cm}p{2.5cm}p{3.5cm}p{4.0cm}}%
\toprule \textbf{Code} & \textbf{Cognitive Function} & \textbf{Meaning /
Operation} & \textbf{Typical Rhetorical Modes Realizing the Function}\\
\midrule C(1) & Observe & Register phenomena and qualitative data. &
Description, Narration, Exemplification, \textbf{Evidence}\\[3pt]%
C(2) & Identify & Distinguish entities or patterns. & Definition, Contrast,
Classification, \textbf{Evidence}\\[3pt]%
C(3) & Compare & Relate attributes or outcomes. & Comparison, Analogy,
Evaluation\\[3pt]%
C(4) & Classify & Categorize into structured groups. & Classification,
Division, Definition\\[3pt]%
C(5) & Abstract & Extract general properties or patterns. & Exemplification,
Exposition, Analogy\\[3pt]%
C(6) & Hypothesize & Formulate possible explanations. & Problem, Cause,
Argumentation\\[3pt]%
C(7) & Model & Represent systems or relations symbolically. & \textbf{Process
Analysis}, Analogy, Exposition\\[3pt]%
C(8) & Infer & Deduce implications or rules. & Cause, Effect,
Argumentation\\[3pt]%
C(9) & Test / Validate & Evaluate hypotheses or models against evidence. &
\textbf{Evidence}, Illustration, Evaluation\\[3pt]%
C(10) & Explain & Provide causal or functional accounts. & Cause and Effect,
Exposition, \textbf{Process Analysis}\\[3pt]%
C(11) & Evaluate & Judge validity or relevance of claims. & Evaluation,
Comparison, Argumentation, \textbf{Persuasion}\\[3pt]%
C(12) & Predict & Anticipate outcomes or states. & Analogy, Cause, Process
Analysis\\[3pt]%
C(13) & Integrate / Synthesize & Combine diverse reasoning into a coherent
whole. & Exposition, Analogy, Synthesis (from Analysis ynthesis pair)\\[3pt]%
C(14) & Reflect / Meta-cognitive Assessment & Assess one's reasoning or
knowledge limits. & Evaluation, Definition, Problem-solution, Exposition,
\textbf{Persuasion}\\[3pt]%
\bottomrule &  &  &
\end{tabular}
\end{table}

\subsubsection{Cognitive Layer to Epistemic Layer}

The epistemic layer is supposed to contain $K_{E}$ epistemic functions,
denoted $E\left(  k\right)  ,k=1,2,\cdots,K_{E}$. It is evident that
\textbf{Teaching / Learning} is one such function, which means pedagogical
transmission of knowledge, with the help of cognitive functions, such as
\textbf{Observe}, \textbf{Identify}, \textbf{Model}, \textbf{Explain},
\textbf{Evaluate}, and \textbf{Reflect / Meta-cognitive Assessment}, from the
C layer. Table~\ref{tab:epistemic_mapping} lists all $K_{E}=8$ epistemic
functions, including the code, function name, meaning or purpose, and typical
cognitive functions in mapping, i.e., cognitive operations realizing each
epistemic purpose.

\begin{table}[ptbh]
\caption{Mapping between Epistemic Functions (E-layer) and Cognitive Functions
(C-layer)}%
\label{tab:epistemic_mapping}
\centering
\begin{tabular}
[c]{@{}p{0.9cm}p{3.2cm}p{4.0cm}p{5.0cm}}%
\toprule \textbf{Code} & \textbf{Epistemic Function} & \textbf{Meaning /
Purpose} & \textbf{Typical Cognitive Functions Realizing the Epistemic
Purpose}\\
\midrule E(1) & Knowledge Formation & Learning and conceptual development. &
Observe, Identify, Classify, Abstract\\[3pt]%
E(2) & Scientific Discovery & Forming and testing new hypotheses. &
Hypothesize, Model, Test / Validate, Infer\\[3pt]%
E(3) & Communication & Dissemination of knowledge through text or discourse. &
Compare, Explain, Integrate / Synthesize, Evaluate\\[3pt]%
E(4) & Teaching / Learning & Pedagogical transmission of knowledge. & Observe,
Identify, Model, Explain, Evaluate, Reflect / Meta-cognitive Assessment\\[3pt]%
E(5) & Problem Solving & Practical or scientific application of knowledge. &
Classify, Hypothesize, Infer, Test / Validate, Evaluate\\[3pt]%
E(6) & Innovation / Design & Creative generation of new systems or methods. &
Model, Abstract, Integrate / Synthesize, Predict\\[3pt]%
E(7) & Evaluation / Decision-Making & Policy or strategic assessment based on
evidence. & Evaluate, Reflect / Meta-cognitive Assessment, Infer\\[3pt]%
E(8) & Policy / Action Implementation & Applying knowledge in real contexts. &
Predict, Explain, Integrate / Synthesize, Evaluate\\[3pt]%
\bottomrule &  &  &
\end{tabular}
\end{table}

Again, the $8$ epistemic functions listed in Table
~\ref{tab:epistemic_mapping} are not exclusive. For some specific applications
we may have more functions.

\subsubsection{An Example in Teaching/Learning}

\begin{figure}[ptbh]
\centering\includegraphics[width=0.85\textwidth]{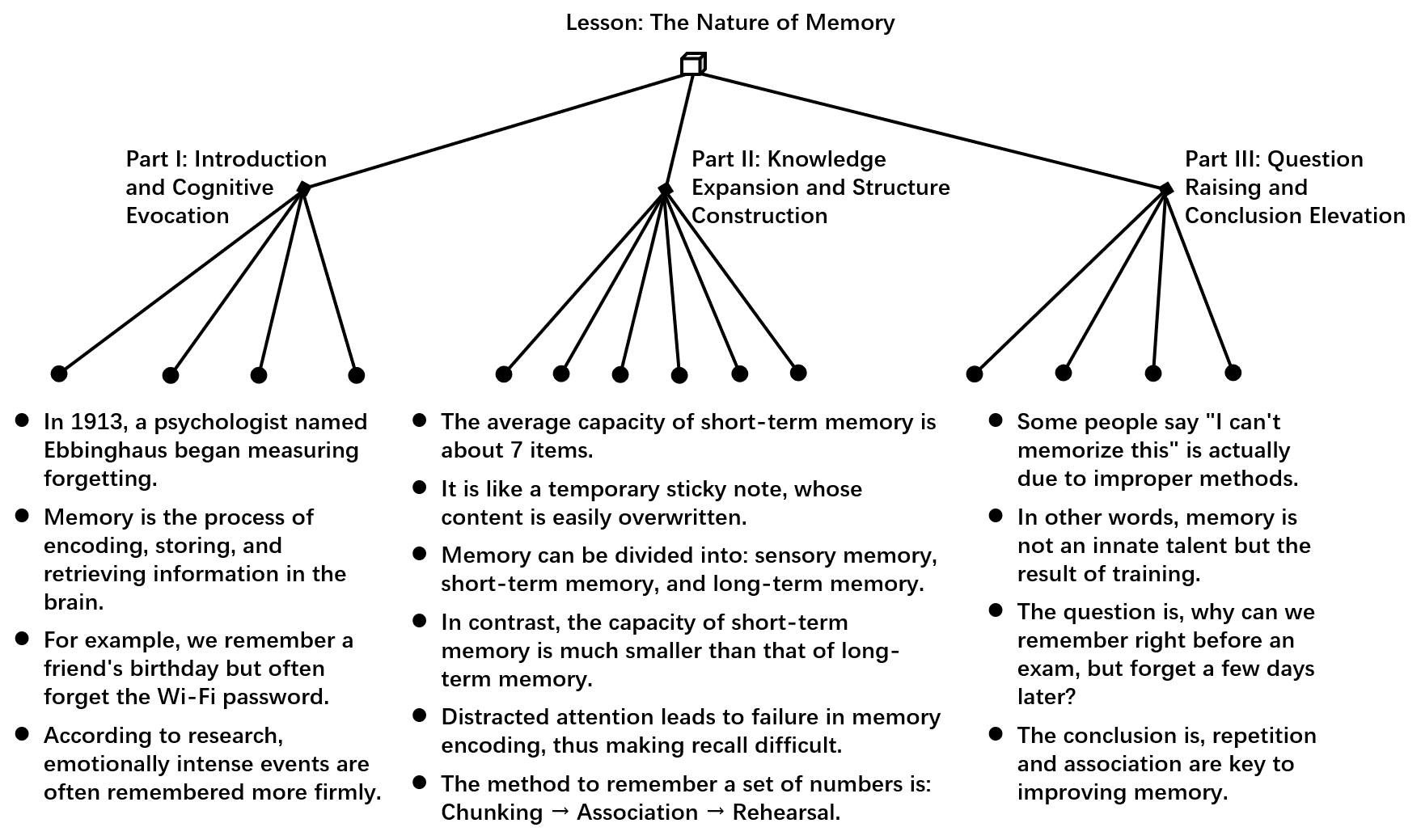}\caption{Composition
of a contents in a lesson}%
\label{fig4-1}%
\end{figure}

\begin{figure}[ptbh]
\centering\includegraphics[width=0.85\textwidth]{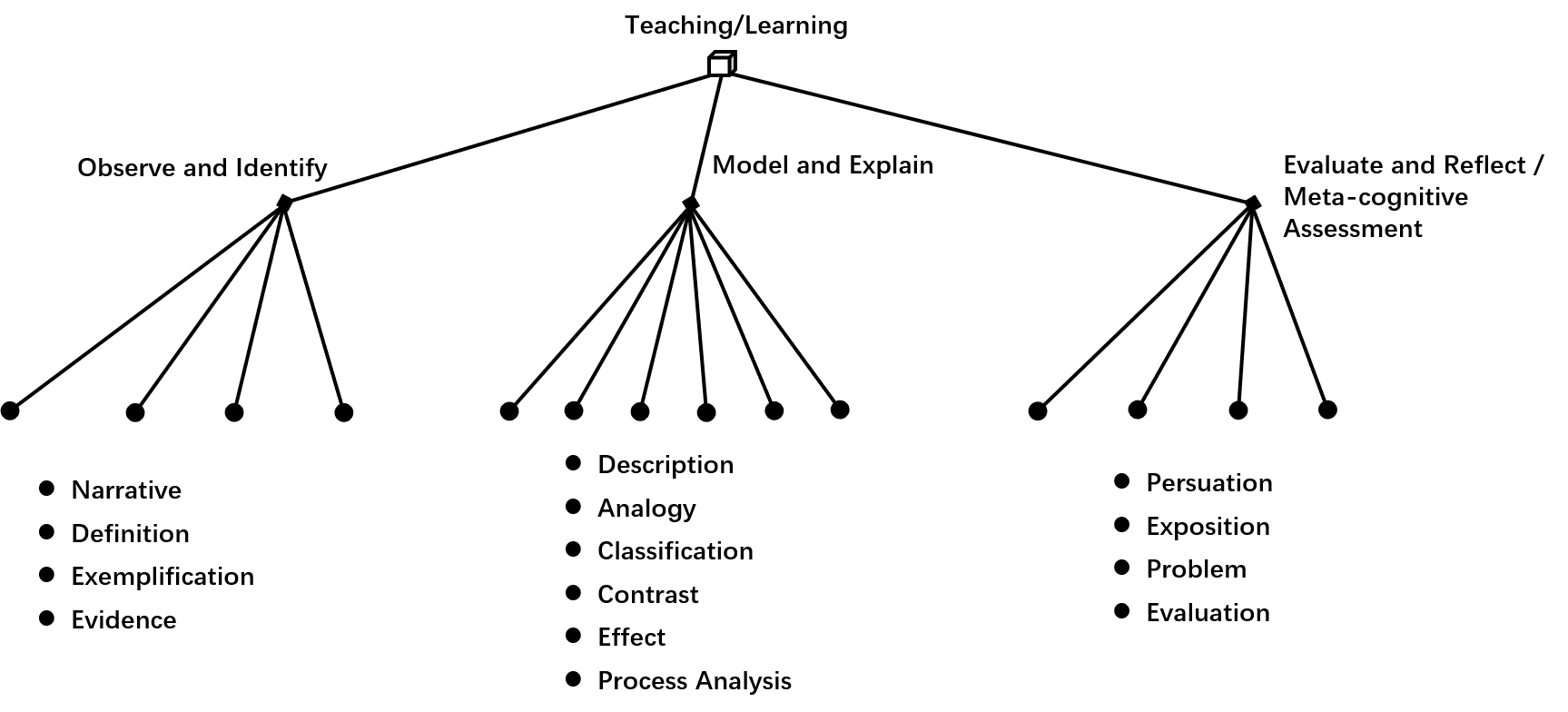}\caption{Correspondence
in three layer mapping from the composition of a contents in a lesson shown in
Figure \ref{fig4-1}}%
\label{fig4-2}%
\end{figure}

As illustration, we take the example of Wu (2025a), which hyperbolized in a
classroom lesson entitled \textquotedblleft The Nature of
Memory\textquotedblright, the teacher\ uses the method of
rhetorical-mode-driven instruction with three layers. The example demonstrates
how strategic sequencing of modes can realize distinct pedagogical functions
through introducing a cognitive layer. Figure \ref{fig4-1} shows the details
of the lesson, and Figure \ref{fig4-2} is the correspongding three layer mapping.

The lesson "The Nature of Memory" corresponds to \textbf{Teaching / Learning}
in the epistemic layer.

Part I (Introduction and Cognitive Evocation) corresponds to the cognitive
functions \textbf{observe} and \textbf{identify}, and uses four rhetorical
modes (narrative, definition, exemplification and evidence) to stimulate
interest and establish topic background.

Part II (Knowledge Expansion and Structure Construction) corresponds to the
cognitive functions \textbf{model} and \textbf{explain}, and uses six
rhetorical modes (description, analogy, classification, contrast, effect,
process analysis) to accomplish quantitative description $\rightarrow$
deconstruct structure$\rightarrow$ establish relationships $\rightarrow$ guide
operational path.

Part III (Question Raising and Conclusion Elevation) corresponds to the
cognitive functions \textbf{evaluate} and \textbf{reflect / meta-cognitive
assessment}, and uses four rhetorical modes (persuation, exposition problem,
evaluation) to summarize causes $\rightarrow$ eliminate cognitive
misconceptions $\rightarrow$ raise common questions $\rightarrow$ elevate into
teaching conclusions.{}

\vspace{0.5em} \noindent The lesson exemplifies a three-tiered orchestration:

\begin{enumerate}
\item Evocation complex (narrative + definition + example + evidence)

\item Construction complex (description + classification + comparison +
cause--effect + process)

\item Reflection complex (argument + clarification + question + evaluation)
\end{enumerate}

Through deliberate sequencing, the instructor activates cognitive curiosity,
builds conceptual structure, and consolidates understanding. The case confirms
that effective pedagogy inherently employs cognitive mapping to bridge
information and cognition. This educational example demonstrates several
theoretical implications.

\begin{itemize}
\item Rhetorical layering as cognitive scaffolding: Each layer of the lesson
corresponds to a cognitive stage---activation, construction, and
reflection---showing that rhetorical orchestration parallels the learning
cycle proposed in educational psychology.

\item Transferability across disciplines: The same compositional logic governs
a physics explanation, a history lecture, or a literature analysis, in the way
that knowledge is built by combining definitional, exemplificatory, causal,
and evaluative operations.
\end{itemize}

\subsubsection{Alternative modes or functions in each layer and refreshed
rhetorical modes for academic function}

The three-layer mapping describes a canonical path - from rhetorical modes
($R$) to cognitive operations ($C$) and then to epistemic purposes ($E$)-
based on abstract functional classification. However, the rhetorical modes and
cognitive-layer functions (Tables~\ref{tab:cognitive_mapping}) and
epistemic-layer functions (Table~\ref{tab:epistemic_mapping}) are \textbf{not
exclusive}. For any specific applications, one may have different modes or functions.

For instance, in some type of academic writing, the rhetorical modes may be
refreshed to generate academic functions through combining a core rhetorical
mode and some supplementart modes (Wu 2025b). The cognitive functions may be
replaced by the following seven ones, each is mapped into some refreshed
rhetorical modes, here called academic functions, in the R-layer.

1. R- layer: information presentation function

This function organizes and presents foundational information of the research
subject - its background, features, structure, and hierarchy. The typical
academic functions are:narration, description, definition, classification,
decomposition, grading, summarization, delineation. Cognitive operation:
perceptual organization $\Rightarrow$ Epistemic purpose: establishing shared
conceptual ground.

2. R- layer: relational reasoning function

This function analyzes relationships among entities to construct explanatory
or theoretical frameworks. The typical academic functions are:
\textit{comparison, contrast, analogy, relational analysis, causal analysis,
induction, synthesis}. Cognitive operation: \emph{analytical linking}
$\Rightarrow$ Epistemic purpose: \emph{constructing relational understanding.}

3. R- layer: process construction function

This function explains sequences, developments, or operational steps, common
in methods and procedures. The typical academic functions are::
\textit{process analysis, procedural description}. Cognitive operation:
\emph{sequential reasoning} $\Rightarrow$ Epistemic purpose: \emph{revealing
operational logic.}

4. R- layer: argumentation support function

This builds and supports authorial claims and reasoning chains. Typical
academic functions are: \textit{exemplification, evidence, argumentation,
persuasion, elaboration, claim-making}. Cognitive operation: \emph{inferential
justification} $\Rightarrow$ Epistemic purpose: \emph{validating
propositions.}

5. R- layer: understanding construction function

This deepens conceptual clarity and connects known with unknown. Typical
academic functions are: \textit{clarification, explanation}. Cognitive
operation: \emph{integrative comprehension} $\Rightarrow$ Epistemic purpose:
\emph{achieving conceptual coherence.}

6. R- layer: interaction construction function\newline This duides the reader
for thinking and invites engagement. Typical academic functions are:
\textit{questioning, answering}. Cognitive operation: \emph{dialogic
coordination} $\Rightarrow$ Epistemic purpose: \emph{co-constructing
understanding.}

7. R- layer: evaluation and reflection function

This assesses value or validity of ideas, methods, or results. Typical
academic functions are: \textit{evaluation, verification, validation}.
Cognitive operation: \emph{critical judgment} $\Rightarrow$ Epistemic purpose:
\emph{determining reliability and significance.}

\subsubsection{Mapping through code mode strenthened by supplementary modes}

As mentioned above, academic functions ($R_{\text{academic}}$) are still in
the R- layer, and represent refreshed modes, each is obtained through
combining a core rhetorical mode (also called dominant mode) and some
supplementart modes (also called auxiliary mode). These refreshed modes are
more closer to cognitive functions.

Formally, each academic function emerges through a special mapping that
combines a core rhetorical mode with several supplementary modes:%

\[
R_{\text{core}}+\sum_{i=1}^{m}R_{\text{sup},i}\Rightarrow R_{\text{academic}}%
\]

An academic function may have the same name of a rhetorical mode, such as
definition, or different name, such as claim.

1. Example one: definition as academic function

When definition is strengthened to be an academic function, then the core mode
is Definition and the supplementary ones are Comparison, Exemplification,
Classification. The \emph{definition complex} expands a simple defining act
into a multidimensional explanation in the following way

\begin{itemize}
\item Core mode: Definition - states the essential meaning (e.g.
\textquotedblleft Memory is the mental process by which information is
encoded, stored, and retrieved.\textquotedblright)

\item Supplementary 1: Comparison - clarifies by opposition or
similarity.(e.g.\textquotedblleft Unlike perception, which involves immediate
sensory input, memory preserves information beyond the present
moment.\textquotedblright)

\item Supplementary 2: Exemplification - grounds meaning in
instance.(e.g.\textquotedblleft For instance, recalling a friend's birthday or
the melody of a song illustrates long-term memory.\textquotedblright)

\item Supplementary 3: Classification - locates the term within a category
system.(e.g.\textquotedblleft Memory can be divided into sensory, short-term,
and long-term types, each serving distinct cognitive
functions.\textquotedblright)
\end{itemize}

Together, these four operations form a higher-order definition mode that
realizes the academic function $(\text{Definition})$, performing
\emph{conceptual structuring} more closer to the cognitive layer and
\emph{establishing academic meaning} at the epistemic layer: 2. Example two:
claim as academic function

For the claim function, the core rhetorical mode is Description or
Cause--Effect, and the supplementary ones are Argumentation or Process
Analysis, Evaluation, Comparison and Evidence. The \emph{claim complex} thus
represents an academic function distinct from its core rhetorical base. It
transforms a mere assertion into a reasoned, validated proposition, in the
following way

\begin{itemize}
\item Core mode: Description / Cause--Effect -illustates a relation (e.g.
\textquotedblleft Theorem~1. In any right triangle, the hypotenuse is longer
than either leg.\textquotedblright)

\item Supplementary 1: Argumentation or Process Analysis - explains why it
holds. (e.g.\textquotedblleft Because the square of the hypotenuse equals the
sum of the squares of the other two sides, its measure must exceed each
leg.\textquotedblright)

\item Supplementary 2: Evaluation- assesses disciplinary importance.
(e.g.\textquotedblleft This theorem forms the foundation of Euclidean geometry
and underpins trigonometric calculation.\textquotedblright)

\item Supplementary 3: Comparison - differentiates it from alternative claims.
(e.g.\textquotedblleft Unlike empirical generalizations, this theorem is
derived deductively rather than from observation.\textquotedblright)

\item Supplementary 4: Evidence - anchors the claim in verification.
(e.g.\textquotedblleft This result is confirmed by Euclid - Elements, Book~I,
Proposition~47.\textquotedblright)
\end{itemize}

When integrated, these operations create a higher-order claim mode realizing
$\text{Claim}$, which performs \emph{inferential justification} cognitively
and \emph{knowledge validation} epistemically.

\section{Conclusion}

This study has sought to reinterpret rhetorical modes as measurable,
generative elements rather than static pedagogical categories. By introducing
non-exclusive four duality-based mode operations (split-unite duality,
forward-backward duality, expansion-reduction duality, orthogonal duality),
the work demonstrates how new rhetorical forms can be derived from canonical
ones, yielding an extended, more symmetrical system of expressive operations.
Quantitative modeling through binomial combinatorics shows that rhetorical
capacity expands exponentially with the number of available modes, and the
notion of a Marginal Rhetorical Bit (MRB) offers a simple measure of
expressive gain per added mode. The rate of rhetorical introduction further
enables the definition of a rhetorical-scalable parameter, connecting
rhetorical growth with cognitive development across educational stages.

To reduce the cognitive complexity in using a growing number of rhetorical
modes, this study also proposed a\ pyramid multilayer mapping framework that
links rhetorical, cognitive, and epistemic layers. The proposed mappings
provide a conceptual bridge between linguistic diversity and knowledge
structure, suggesting that rhetorical systems can be both expansive and
cognitively economical.

While the \emph{pyramid multilayer mapping} defines a conceptual taxonomy
linking rhetorical, cognitive, and epistemic layers in a hierarchical manner
and offers a structural ontology of functions (e.g., $R\rightarrow
C\rightarrow E$), however, the modes or functions in each layer should not be
considered as fixed. For instance, for academic writing, the R layer could use
refreshed rhetorical mode (academic function) through through compositional
orchestration, i.e., through combining a core mode and some supplementary
modes. The same may be true for cognitive layer and epistemic layer.

Several limitations should be acknowledged. For instance, there lacks a formal
way to \ evaluate whether a generated rhetorical mode can be used as a
rhetorical mode. The mappings among layers, while conceptually clear, could
vary across disciplines and languages, requiring further validation. Moreover,
the operator calculus for mode generation is presented schematically and would
benefit from formalization in symbolic or computational terms.

Even with these constraints, the results indicate that rhetorical modes can be
studied quantitatively and modeled systematically. Such an approach could
inform writing pedagogy, academic writing, and computational language models.
Future work could explore how AI systems might apply layered rhetorical
reasoning in discourse generation. In this sense, the present study offers an
initial step toward integrating rhetorical mode theory with quantitative and
computational perspectives in a careful, incremental manner.\newline

\textbf{Acknowledgements} This work draws upon insights accumulated over six
years of teaching academic writing to undergraduate and graduate students,
whose thoughtful questions and reflections continually inspired the
development of this study. The author made limited use of AI-powered language
tools to assist with linguistic refinement and the formulation of illustrative
examples. Post-doc student Wang MM and PhD student Wang XY helped to prepare
the latex file and some figures.


\subsection*{References}

%\ibliography{refs}
%%\bibliography{sn-bibliography}% common bib file
Aristotle. (2007). \textit{On Rhetoric: A Theory of Civic Discourse} (G.~A.
Kennedy, Trans.). Oxford University Press. (Originally composed 4th century BCE)

Axelrod, R.~B., \& Cooper, C.~R. (2020). \textit{The St.~Martin's Guide to
Writing} (13th~ed.). Bedford/St.~Martin's.

Baayen, R. H. (2001). Word frequency distributions. Springer

Bain, A. (1866). English composition and rhetoric. D. Appleton and Company.

Barnet, S., \& Bedau, H. (2014). \textit{Critical Thinking, Reading, and
Writing: A Brief Guide to Argument} (8th~ed.). Bedford/St.~Martin`s Press.

Bartholomae, D., \& Petrosky, A. (2008). \textit{Ways of Reading: An Anthology
for Writers} (9th~ed.). Bedford/St.~Martin's.

Bazerman, C. (1988). \textit{Shaping Written Knowledge: The Genre and Activity
of the Experimental Article in Science.} University of Wisconsin Press.

Beaufort, A. (2007). \textit{College Writing and Beyond: A New Framework for
University Writing Instruction.} Utah State University Press.

Behrens, L., \& Rosen, L.~J. (2010). \textit{Writing and Reading Across the
Curriculum} (11th~ed.). Longman.

Bettencourt, L. M. A., Lobo, J., Helbing, D., Kunert, C., \& West, G. B.
(2007). Growth, innovation, scaling, and the pace of life in cities.
Proceedings of the National Academy of Sciences, 104(17), 7301--7306.

Booth, W.~C., Colomb, G.~G., \& Williams, J.~M. (2008). \textit{The Craft of
Research} (3rd~ed.). University of Chicago Press.

Brown, M. D., \& Hagerty, M. E. (1917).(1917). The measurement of improvement
in English composition. English Journal, 6(8), 515--527.

Connors, R. J. (1997). Composition-Rhetoric: Backgrounds, theory, and
pedagogy. University of Pittsburgh Press.

Corbett, E.~P.~J., \& Connors, R.~J. (1999). \textit{Classical Rhetoric for
the Modern Student} (4th~ed.). Oxford University Press.

Cover, T. M., \& Thomas, J. A. (2006). Elements of information theory (2nd
ed.). Wiley-Interscience

Flavell, J.~H. (1979). Metacognition and cognitive monitoring: A new area of
cognitive-developmental inquiry. \textit{American Psychologist, 34}(10),
906--911. \url{https://doi.org/10.1037/0003-066X.34.10.906}

Flower, L., \& Hayes, J.~R. (1981). A cognitive process theory of writing.
\textit{College Composition and Communication, 32}(4), 365--387. \url{https://doi.org/10.2307/356600}

Graff, G., \& Birkenstein, C. (2021). \textit{They Say / I Say: The Moves That
Matter in Academic Writing} (5th~ed.). W.~W. Norton.

Hacker, D., \& Sommers, N. (2020). \textit{A Writer's Reference} (10th~ed.). Bedford/St.~Martin's.

Herrick, J.~A. (2020). \textit{The History and Theory of Rhetoric: An
Introduction} (7th~ed.). Routledge.

Hill, A. S. (1895). The principles of rhetoric. Harper \& Brothers.

Hoey, M. (1983). \textit{On the Surface of Discourse}. Allen \& Unwin.

Hyland, K. (2004). \textit{Genre and Second Language Writing.} University of
Michigan Press.

Jordan, R.~R. (1999). \textit{Academic Writing Course: Study Skills in
English} (3rd~ed.). Longman.

Kane, T.~S. (2000). \textit{The Oxford Essential Guide to Writing.} Oxford
University Press.

Knessl, C. (1998). Integral representations and asymptotic expansions for
Shannon and Renyi entropies. Applied Mathematics Letters, 11(2), 69--74.

Kirszner, L.~G., \& Mandell, S.~R. (1986). \textit{Patterns for College
Writing: A Rhetorical Reader and Guide.} St.~Martin's Press.

Lakoff, G., \& Johnson, M. (1980). \textit{Metaphors We Live By.} University
of Chicago Press.

Langacker, R.~W. (1987). \textit{Foundations of Cognitive Grammar} (Vol.~1).
Stanford University Press.

Langan, J. (2013). \textit{College Writing Skills with Readings} (9th~ed.). McGraw--Hill.

Lotman, Y. M. (1990). Universe of the mind: A semiotic theory of culture.
Indiana University Press.

Lunsford, A.~A. (2015). \textit{The St.~Martin's Handbook} (8th~ed.). Bedford/St.~Martin's.

Lunsford, A.~A., Ruszkiewicz, J.~J., \& Walters, K. (2021).
\textit{Everything's an Argument} (9th~ed.). Bedford/St.~Martin's.

McQuade, D., \& Atwan, R. (1998). \textit{The Writer's Presence: A Pool of
Readings.} Bedford/St.~Martin's.

Nadell, J., Langan, J., \& Coxwell-Teague, D. (2019). \textit{The Longman
Reader} (12th~ed.). Pearson.

Newell, A., \& Rosenbloom, P. S. (1981). Mechanisms of skill acquisition and
the law of practice. In J. R. Anderson (Ed.), Cognitive skills and their
acquisition (pp. 1--55). Lawrence Erlbaum Associates.

Oshima, A., \& Hogue, A. (2007). \textit{Writing Academic English} (4th~ed.).
Pearson Longman.

Piaget, J. (1952). The origins of intelligence in children. International
Universities Press.

Reed, C., \& Norman, T. (Eds.). (2004). Argumentation machines: New frontiers
in argument and computation. Springer.

Rosenwasser, D., \& Stephen, J. (2019). \textit{Writing Analytically}
(8th~ed.). Cengage Learning.

R{\o }rstad, K., \& Aksnes, D.~W. (2015). Publication rate expressed by age,
gender, and academic position: A large-scale analysis of Norwegian academic
staff. \textit{Journal of Informetrics, 9}(2), 255--266. \url{https://doi.org/10.1016/j.joi.2015.02.003}

Shannon, C. E. (1948). A mathematical theory of communication. The Bell System
Technical Journal, 27(3), 379--423.

Smalley, R.~L., Ruetten, M.~K., \& Kozyrev, J.~R. (2011). \textit{Refining
Composition Skills: Rhetoric and Grammar for ESL Students} (6th~ed.). Heinle Cengage.

Simonton, D. K. (2004). Creativity in science: Chance, logic, genius, and
Zeitgeist. Cambridge University Press.

Sollaci, L. B., \& Pereira, M. G. (2004). The introduction, methods, results,
and discussion (IMRAD) structure: A fifty-year survey.\textit{\ Journal of the
Medical Library Association}, 92(3), 364--371.

Wu, Z.~N. (2025a). \textit{How to Navigate in University: Four Foundational
Thinking Skills Beyond Knowledge} (in Chinese). Shanxi Normal University Press.

Wu, Z.~N. (2025b). \textit{Four Layer Progressive Writing for Thesis and
Dissertation } (in Chinese). Shanxi Normal University Press. In Press.

Young, R. E. (1978). Paradigms and problems: Needed research in rhetorical
invention. In C. R. Cooper \& L. Odell (Eds.), Research on composing: Points
of departure (pp. 29--47). National Council of Teachers of English.

Zipf, G. K. (1949). Human behavior and the principle of least effort.
Addison--Wesley.

\end{document}